\documentclass[11pt]{article}

\usepackage[preprint]{acl}

\usepackage{times}
\usepackage{latexsym}
\usepackage{placeins}
\usepackage[T1]{fontenc}

\usepackage[utf8]{inputenc}

\usepackage{microtype}

\usepackage{inconsolata}

\usepackage{graphicx}

\usepackage{bm}
\usepackage{tabularx}
\usepackage{array}
\usepackage{amsmath,amssymb}
\usepackage{booktabs}
\usepackage{tablefootnote}

\title{From Token Correction to Claim Admission: \\ Training-Free Hallucination Mitigation for Large Vision-Language Models}

\author{
  Hyeonggeun Han \\
  Independent Researcher \\
  \texttt{hygnhan@gmail.com}
  \And
  Taehyun Cho \\
  Vector Institute \\
  Fields Institute \\
  Seoul National University \\
  \texttt{talium@cml.snu.ac.kr}
  \And
  Sangwoo Hong \\
  Konkuk University \\
  \texttt{swhong06@konkuk.ac.kr}
}

\begin{document}
\maketitle
\begin{abstract}
Large vision-language models often exhibit object hallucination, generating claims that are not supported by the input image. To mitigate this issue, we propose Evidence-Guided Visual Claim Admission (EGVCA), a training-free framework that admits a semantic claim only when it receives sufficient image-dependent support. EGVCA measures this support by comparing the claim likelihood under the observed image and a visual-null input. For captioning, it controls semantic expansion without altering token generation. For binary visual question answering, it combines atomic visual support with the image-induced change in the model's native Yes--No margin. Experiments on CHAIR, POPE, and MME across three LVLM backbones show that EGVCA substantially reduces object hallucination while preserving caption quality. It achieves the highest POPE F1 in eight of nine backbone--sampling settings and improves the native MME total score on all three backbones. These results demonstrate the effectiveness of claim-level admission as a simple alternative to token-level distribution correction.
\end{abstract}

\section{Introduction}

Large Vision-Language Models (LVLMs) have demonstrated remarkable capabilities in multimodal understanding and generation, enabling a broad range of applications that require interaction with visual environments. Despite this progress, their reliability remains fundamentally challenged by object hallucination (OH), where generated responses contain object-centric information that is inconsistent with the visual input \cite{li2023pope,wang2023amber,wang2026after}. Such hallucinations can manifest in various forms, including mentioning objects that are not present, describing incorrect attributes or properties of visible objects, or asserting relationships between objects that are not supported by the image. These errors can remain linguistically plausible even when they are visually ungrounded, making them particularly difficult to detect from the generated response alone. As LVLMs are increasingly expected to provide faithful interpretations of visual scenes, mitigating OH is therefore essential for improving their reliability and enabling trustworthy real-world deployment.

\begin{figure*}[t]
    \centering
    \includegraphics[width=\textwidth]{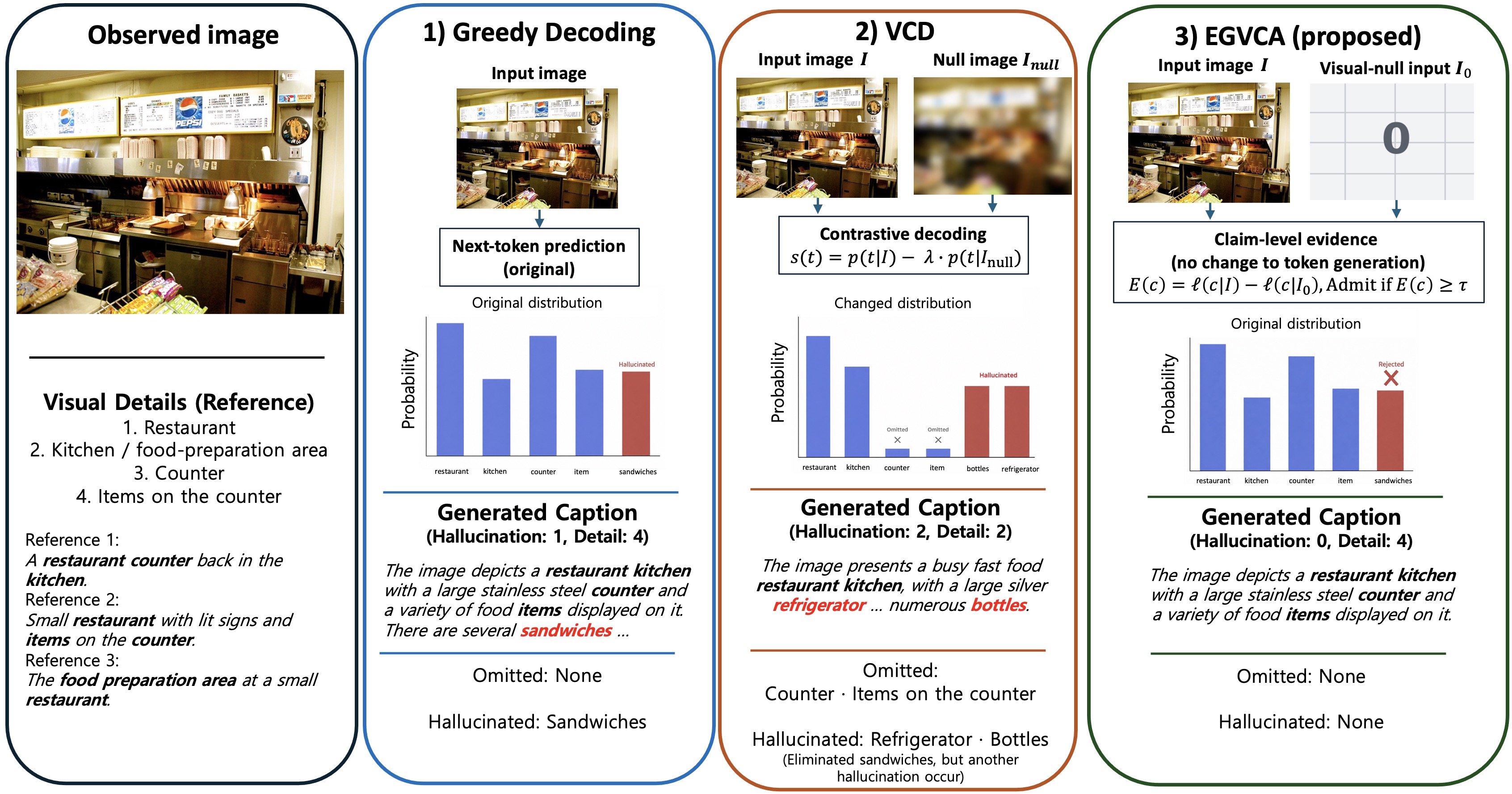}
    \caption{Comparison of greedy decoding, VCD, and EGVCA. Unlike VCD, which modifies next-token predictions, EGVCA uses original-versus-visual-null evidence to reject unsupported claims without altering candidate generation.}
    \label{fig:egvca_overview}
\end{figure*}
Numerous approaches have been proposed to reduce hallucinations through additional training, external supervision, representation intervention, and decoding-time correction. Among them, training-free decoding methods have received particular attention because they can be directly applied to pretrained LVLMs without additional training data or parameter updates. A major line of work improves visual grounding by contrasting predictions obtained under different visual or conditional contexts \cite{leng2024vcd,favero2024m3id}, while another exploits internal decoding signals, such as attention patterns, intermediate-layer predictions, or visual-token and layer-wise information, to suppress hallucination-prone outputs \cite{huang2024opera,wang2025deco,huo2025sid,liang2025mole}. More recent approaches further suggest that correcting generation selectively according to the visual support of emerging content can avoid unnecessary intervention \cite{yang2026vgd}. Despite their effectiveness, existing methods primarily mitigate hallucination by modifying token distributions or intervening on particular internal signals during decoding. Such correction can suppress hallucination-prone generation, but modifying the generation process itself may also alter visual information that is already well grounded. As a result, the model may favor safer but less semantically specific responses. We therefore take a different perspective. Rather than directly correcting how a claim is generated, we first ask whether the semantic claim itself has sufficient visual evidence to be admitted into the response. This separation allows hallucination mitigation to focus on unsupported semantic commitments while leaving grounded generation otherwise unchanged.

Motivated by this perspective, we propose
\textbf{Evidence-Guided Visual Claim Admission (EGVCA)}, a training-free
framework that formulates hallucination mitigation as visual claim
admission. As illustrated in Figure~\ref{fig:egvca_overview}, EGVCA is
based on a simple principle: a visual claim should be incorporated into
the response only when its plausibility can be meaningfully attributable to
the observed image. We estimate this support by comparing the normalized
likelihood of the same claim under the observed image and a visual-null
input, implemented as an all-zero preprocessed visual tensor. This
counterfactual comparison isolates image-dependent evidence from
plausibility induced by linguistic context alone.

Unlike contrastive decoding methods that directly modify the next-token
distribution, EGVCA interprets the resulting contrast as evidence for a
semantic claim. It then uses this evidence to regulate whether the model
commits to that claim, while leaving the underlying candidate generation
unchanged. This separation between claim-level evidence estimation and
claim admission provides a unified mechanism for suppressing visually
unsupported content while retaining supported semantic information.
Experiments across three LVLM backbones on CHAIR, POPE, and MME show that EGVCA consistently reduces object hallucination while maintaining
competitive generation quality.

\section{Related Work}
\paragraph{Large Vision-Language Models.}
LVLMs have rapidly advanced by integrating visual encoders with large language models, achieving strong performance across a wide range of multimodal understanding and generation tasks \cite{li2023blip2,liu2023llava,zhu2024minigpt4,ye2024mplugowl2,bai2023qwenvl}. Recent progress in large-scale multimodal pretraining and visual instruction tuning has further improved their ability to follow natural-language instructions and reason over complex visual content. As a result, LVLMs have become a central foundation for general-purpose multimodal systems. Despite these advances, imperfect visual grounding remains a persistent limitation, causing LVLMs to generate semantically plausible content that is inconsistent with the visual input.

\paragraph{Object Hallucination in LVLMs.}
Object hallucination (OH) refers to the generation of object-centric content that is not faithfully grounded in the given image \cite{rohrbach2018object,li2023pope}. It can appear as incorrect object identification as well as erroneous attributes, quantities, or relations associated with visual entities \cite{wang2023amber}. Prior studies have linked OH to biased object co-occurrences, strong language priors, insufficient vision-language alignment, and the diminishing influence of visual information during autoregressive generation \cite{li2023pope,leng2024vcd,favero2024m3id}. To address these issues, various mitigation strategies have been explored, with training-free inference-time methods receiving particular attention for improving visual faithfulness without modifying the underlying LVLM.

\paragraph{Training-Free Decoding for Hallucination Mitigation.}
Training-free decoding provides an efficient means of mitigating
hallucinations without retraining LVLMs. Many approaches derive corrective
signals from the predictions or internal computations of a frozen model.
VCD contrasts predictions under the original and distorted visual inputs
to suppress language priors, while OPERA penalizes over-trusted attention
patterns during decoding \citep{leng2024vcd,huang2024opera}. DeCo uses
visual information preserved in intermediate layers to correct next-token
predictions, and MoLE combines predictions from multiple layer-wise experts
to improve visual grounding \citep{wang2025deco,liang2025mole}. Recent
analyses further show that reliance on linguistic context tends to increase
as generation proceeds, while visual dependency can weaken during later
decoding stages \citep{min-etal-2025-mitigating,anand2026crops}. This
tendency motivates explicit control over the visual support of later
generated content. The use of such internal signals is also connected to broader research that seeks to understand
and control model behavior through interpretable internal features,
mechanistic circuits, role-aware information flow, confidence-aware fusion,
and evidence-grounded structured representations
\citep{huben2024sparse,lee2026mechanistic,lee2026hallucination,
YeoChan2026Adaptiveorthogonal,lee-etal-2025-large}. EGVCA shares the broader goal of deriving
informative signals from a frozen model, but differs in how these signals are
used. Rather than directly modifying next-token predictions, EGVCA interprets
the original-versus-null likelihood contrast as evidence for a semantic claim
and uses it to determine whether that claim should be admitted.



\section{Claim-Level View of Object Hallucination}
\label{sec:problem_formulation}

Given an input image $I$ and a textual instruction $x$, a pretrained LVLM parameterized by $\theta$ generates a response $y$ according to $p_\theta(y\mid I,x)$. Rather than viewing hallucination solely as an error at the token level, we consider the semantic commitments expressed in the response. Specifically, we define $y$ as a sequence of semantic claims $y=(c_1,\ldots,c_T)$, where each $c_t$ may describe the existence of an object, an attribute of a visible object, or a relation among visual entities. Letting $h_t=(c_1,\ldots,c_{t-1})$ denote the preceding claims, with $h_1$ being empty, we express the response likelihood as
\begin{equation}
p_\theta(y\mid I,x) = \prod_{t=1}^{T} p_\theta(c_t\mid I,x,h_t).
\label{eq:problem_claim_factorization}
\end{equation}
Under this formulation, object hallucination arises when the model introduces a semantic commitment that is not sufficiently grounded in the observed image.

A central difficulty is that the likelihood of a claim does not by itself indicate whether the claim is visually grounded. A claim may receive high probability because it is linguistically plausible or favored by the model's learned priors even when the image provides little evidence for it. We therefore distinguish the overall plausibility of a claim from the extent to which that plausibility increases in the presence of the observed image. To this end, we consider a visual-null counterpart $I_0$ of the input image $I$, under which task-relevant visual information is removed. This motivates comparing the likelihood of the same claim under $I$ and $I_0$ through the log-likelihood contrast
\begin{equation}
\log p_\theta(c_t\mid I,x,h_t) - \log p_\theta(c_t\mid I_0,x,h_t).
\label{eq:problem_visual_contrast}
\end{equation}
A positive contrast indicates that the observed image increases the model-assigned likelihood of the claim relative to the visual-null input. This comparison provides a basis for assessing image-dependent evidence beyond the claim's overall plausibility.

Crucially, we do not formulate this image-dependent likelihood difference as a corrective signal for directly modifying next-token probabilities. Instead, we treat it as evidence for deciding whether a semantic commitment should be admitted into the model response. This shifts the target of intervention from token generation itself to the model's commitment to semantic content: generation remains unchanged for content that is sufficiently grounded, while visually unsupported commitments can be regulated at the admission stage. We consider this principle in both open-ended generation and binary visual question answering. Although the form of the response and the corresponding admission mechanism differ between the two settings, both instantiate the same underlying problem of determining whether a semantic commitment is sufficiently attributable to the observed image. Section~\ref{sec:method} describes how this likelihood comparison is operationalized as evidence and how claim admission is implemented in each setting.

\section{Evidence-Guided Visual Claim Admission}
\label{sec:method}

Building on the distinction between claim plausibility and image-dependent evidence in Section~\ref{sec:problem_formulation}, we propose Evidence-Guided Visual Claim Admission (EGVCA), a training-free framework for regulating semantic commitments in LVLM responses. EGVCA uses the likelihood contrast in Eq.~\ref{eq:problem_visual_contrast} to estimate visual evidence and guide claim admission. We first define a claim-level visual evidence score and specify the visual-null input used to compute it. We then describe how this evidence determines the retained response prefix in open-ended generation and adjusts the final Yes--No decision in binary visual question answering.

\subsection{Image-Attributable Visual Evidence}
\label{sec:egvca-evidence}

EGVCA normalizes the claim-level likelihood contrast by token count to reduce its dependence on claim length. Using the preceding context $h_t$ defined in Section~\ref{sec:problem_formulation}, we compute the average token log-likelihood of a claim as
\begin{equation}
\ell_\theta(c_t \mid I,x,h_t)
=
\frac{
\sum_{j=1}^{|c_t|}
\log p_\theta(c_{t,j} \mid I,x,h_t,c_{t,<j})
}{
|c_t|
},
\label{eq:claim_likelihood}
\end{equation}
where $|c_t|$ is the number of tokens in $c_t$, $c_{t,j}$ denotes its $j$-th token, and $c_{t,<j}$ denotes its preceding tokens. Averaging places claims of different lengths on a per-token scale. We then define the image-attributable evidence as
\begin{equation}
E(c_t)
=
\ell_\theta(c_t \mid I,x,h_t)
-
\ell_\theta(c_t \mid I_0,x,h_t).
\label{eq:claim_evidence}
\end{equation}
Thus, $E(c_t)$ is the log-likelihood contrast in Eq.~\ref{eq:problem_visual_contrast} divided by $|c_t|$.

The visual-null input $I_0$ is constructed by replacing the backbone-specific preprocessed visual input with a zero-valued tensor of the same shape. This construction removes the image's scene content without substituting another scene or introducing a randomly generated visual pattern. It therefore provides a deterministic reference for measuring the change in claim likelihood induced by the observed image. Accordingly, we interpret $E(c_t)$ as evidence relative to this reference, rather than as an absolute measure of visual correctness.

A positive $E(c_t)$ indicates that the observed image increases the claim's average token log-likelihood relative to $I_0$. A value near zero indicates little change, whereas a negative value indicates that the claim is less likely under the observed image than under the reference. EGVCA uses this score to evaluate claim admission, without feeding it back into the next-token distribution.

\subsection{Claim Admission for Open-Ended Generation}
\label{sec:egvca-open}

For open-ended generation, EGVCA applies evidence-based admission to a candidate response produced by the LVLM's native decoding strategy. We partition this response into the claim sequence described in Section~\ref{sec:problem_formulation} and evaluate the claims in their original order. Each claim is defined as a sentence-like span separated by terminal punctuation. Admission retains a prefix of this sequence without rewriting individual claims.

The first claim $c_1$ is retained without an evidence check to prevent an empty response. Each subsequent claim is evaluated using $E(c_t)$ conditioned on its preceding context $h_t$ and admitted according to
\begin{equation}
\mathrm{Admit}(c_t)
=
\mathrm{I}\left[E(c_t)\geq\tau\right],
\qquad t\geq2,
\label{eq:egvca-open-admission}
\end{equation}
where $\mathrm{I}[\cdot]$ is the indicator function and $\tau$ is the minimum evidence required to retain an additional claim. This rule uses image-attributable evidence to determine whether the response should extend beyond the claims already retained.

Admission stops at the first claim that fails the criterion. If $c_t$ is the first rejected claim, EGVCA discards it together with all subsequent claims and returns
\begin{equation}
\hat{y}
=
(c_1,\ldots,c_{t-1}).
\label{eq:egvca-open-stopping}
\end{equation}
If no claim is rejected, the full candidate response is retained. Stopping at the first rejection preserves the preceding context under which each retained claim was generated and evaluated. It also avoids retaining later claims that may depend on rejected content. The threshold $\tau$ controls how much of the response is admitted beyond the initial claim. Increasing $\tau$ makes admission stricter, while decreasing it allows a longer prefix to be retained.

\subsection{Evidence-Guided Claim Admission for Binary VQA}
\label{sec:egvca-binary}

In binary visual question answering, claim admission concerns the proposition expressed by a question $q$. Answering $\mathrm{Yes}$ affirms this proposition, whereas answering $\mathrm{No}$ rejects it. EGVCA combines evidence about the question's visual components with evidence about the complete proposition, then uses the combined score to adjust the model's final decision.

Component-level evidence measures how strongly the image increases the likelihood of the visual content mentioned in the question. We decompose $q$ into $K$ visual components $a_1,\ldots,a_K$, corresponding to objects, attributes, or relations involved in the proposition. Visual components are extracted using a rule-based parser with predefined lexical variants, without requiring a trained semantic parser. Each component is scored under a fixed neutral visual-description context $x_{\mathrm{neu}}$, which requests visual description without posing a Yes--No question. This context allows us to evaluate the component's textual content outside the original binary question framing. For each component, we compute
\begin{equation}
e(a_k)
=
\ell_\theta(a_k \mid I,x_{\mathrm{neu}})
-
\ell_\theta(a_k \mid I_0,x_{\mathrm{neu}}),
\label{eq:egvca-binary-component-evidence}
\end{equation}
using the token-averaged log-likelihood in Eq.~\ref{eq:claim_likelihood}, with no preceding claim context. When multiple equivalent textual forms are considered for a component, we use the form with the largest evidence score. We aggregate the component scores as
\begin{equation}
V(q)
=
\frac{1}{K}
\sum_{k=1}^{K} e(a_k).
\label{eq:egvca-binary-visual-evidence}
\end{equation}
Averaging reduces the dependence of the aggregate score on the number of components.

Component-level evidence alone does not establish whether the complete proposition is supported. For example, evidence for a dog, a couch, and a sitting action does not necessarily establish that the dog is sitting on the couch. EGVCA therefore includes a second signal that measures how the image changes the model's preference for affirming the question as a whole. We compute the native Yes--No margin under the observed image as
\begin{equation}
m_I(q)
=
\log p_\theta(\mathrm{Yes}\mid I,q)
-
\log p_\theta(\mathrm{No}\mid I,q),
\label{eq:egvca-binary-native-margin}
\end{equation}
and the corresponding margin under the visual-null input as
\begin{equation}
m_0(q)
=
\log p_\theta(\mathrm{Yes}\mid I_0,q)
-
\log p_\theta(\mathrm{No}\mid I_0,q).
\label{eq:egvca-binary-null-margin}
\end{equation}
The proposition-level evidence is their difference,
\begin{equation}
M(q)
=
m_I(q)-m_0(q).
\label{eq:egvca-binary-margin-evidence}
\end{equation}
A positive $M(q)$ indicates that the observed image shifts the model's relative preference toward $\mathrm{Yes}$, whereas a negative value indicates a shift toward $\mathrm{No}$. Unlike $V(q)$, this score evaluates the image's effect on the decision for the complete question.

EGVCA combines these complementary signals through an equally weighted joint evidence score,
\begin{equation}
J(q)
=
\frac{1}{2}V(q)
+
\frac{1}{2}M(q).
\label{eq:egvca-binary-joint-evidence}
\end{equation}
Here, $V(q)$ contributes evidence for the visual components, while $M(q)$ contributes evidence for affirming the complete proposition. The joint score adjusts the native decision margin through
\begin{equation}
m'(q)
=
m_I(q)
+
\lambda\left(J(q)-\tau\right),
\label{eq:egvca-binary-admission-score}
\end{equation}
where $\lambda\geq0$ controls the strength of the adjustment and $\tau$ is the evidence reference level for binary VQA. Evidence above $\tau$ increases the margin toward $\mathrm{Yes}$, while evidence below $\tau$ decreases it. Unlike the hard threshold used for open-ended generation, $\tau$ is not a standalone admission condition. The final decision depends jointly on the native margin and the evidence adjustment.

The final answer is
\begin{equation}
\hat{a}
=
\begin{cases}
\mathrm{Yes}, & m'(q)>0,\\
\mathrm{No}, & m'(q)\leq0.
\end{cases}
\label{eq:egvca-binary-admission}
\end{equation}
This rule applies the evidence adjustment at the final answer-selection stage, leaving the LVLM probabilities used to compute the scores unchanged. Both response settings therefore use visual evidence to regulate semantic commitment, through prefix admission in open-ended generation and decision-level adjustment in binary VQA.

\section{Experiments}
\label{sec:experiments}

\subsection{Experimental Setup}
\label{sec:experimental_setup}

\paragraph{Models.}
We evaluate EGVCA on three pretrained LVLM backbones, LLaVA-1.5-7B~\cite{liu2024llava15}, MiniGPT-4 with Vicuna-13B~\cite{zhu2024minigpt4}, and mPLUG-Owl2-7B~\cite{ye2024mplugowl2}. All model parameters remain frozen throughout evaluation.

\begin{table*}[t]
\centering
\small
\setlength{\tabcolsep}{3.5pt}
\renewcommand{\arraystretch}{1.05}
\caption{CHAIR-500 comparison across three LVLM backbones.}
\label{tab:chair_results}
\resizebox{0.8\textwidth}{!}{%
\begin{tabular}{lrrrrrr}
\toprule[1pt]
& \multicolumn{2}{c}{LLaVA-1.5}
& \multicolumn{2}{c}{MiniGPT-4}
& \multicolumn{2}{c}{mPLUG-Owl2} \\
\cmidrule(lr){2-3}
\cmidrule(lr){4-5}
\cmidrule(lr){6-7}
Method
& $\mathrm{CHAIR}_{S}\downarrow$
& $\mathrm{CHAIR}_{I}\downarrow$
& $\mathrm{CHAIR}_{S}\downarrow$
& $\mathrm{CHAIR}_{I}\downarrow$
& $\mathrm{CHAIR}_{S}\downarrow$
& $\mathrm{CHAIR}_{I}\downarrow$ \\
\midrule
Greedy
& $21.93 \pm 1.79$
& $7.07 \pm 0.59$
& $20.33 \pm 0.52$
& $7.26 \pm 0.39$
& $21.47 \pm 1.00$
& $7.12 \pm 0.39$ \\

Beam
& $18.73 \pm 2.49$
& $6.23 \pm 0.37$
& $20.00 \pm 1.88$
& $7.20 \pm 0.48$
& $19.27 \pm 1.59$
& $6.61 \pm 0.39$ \\

VCD
& $23.87 \pm 0.25$
& $7.91 \pm 0.27$
& $23.73 \pm 0.90$
& $9.02 \pm 0.07$
& $25.40 \pm 2.32$
& $8.77 \pm 0.73$ \\

OPERA
& $18.00 \pm 1.02$
& \underline{$6.02 \pm 0.19$}
& $17.80 \pm 1.57$
& $7.08 \pm 0.35$
& $17.73 \pm 0.74$
& \underline{$6.21 \pm 0.16$} \\

DoLa
& $17.27 \pm 1.51$
& $8.72 \pm 2.08$
& \underline{$15.07 \pm 1.89$}
& $7.08 \pm 0.56$
& $18.87 \pm 1.23$
& $6.44 \pm 0.23$ \\

DeCo
& \underline{$16.20 \pm 1.34$}
& $6.16 \pm 0.65$
& $17.80 \pm 1.31$
& $6.75 \pm 0.46$
& \underline{$17.33 \pm 1.79$}
& $7.33 \pm 1.07$ \\

SID
& $22.13 \pm 0.98$
& $6.50 \pm 0.31$
& $22.93 \pm 2.66$
& $8.65 \pm 0.89$
& $25.40 \pm 1.85$
& $8.68 \pm 0.46$ \\

MoLE
& $18.13 \pm 1.31$
& $6.51 \pm 0.59$
& $17.80 \pm 1.28$
& $6.50 \pm 0.28$
& $18.53 \pm 0.25$
& \underline{$6.21 \pm 0.18$} \\

HALC
& $23.07 \pm 1.75$
& $8.52 \pm 0.62$
& $17.93 \pm 1.15$
& \underline{$6.31 \pm 0.61$}
& $20.00 \pm 1.84$
& $6.83 \pm 0.33$ \\
\midrule
{EGVCA}
& \textbf{10.87 $\pm$ 1.43}
& \textbf{5.77 $\pm$ 0.61}
& \textbf{12.73 $\pm$ 0.66}
& \textbf{6.19 $\pm$ 0.11}
& \textbf{12.60 $\pm$ 1.61}
& \textbf{5.63 $\pm$ 0.70} \\
\bottomrule[1pt]
\end{tabular}}
\end{table*}

\paragraph{Benchmarks and Evaluation Metrics.}
We evaluate object hallucination in open-ended generation using CHAIR~\cite{rohrbach2018object} on 500 images sampled from the MS COCO 2014 validation set~\cite{lin2014coco} per run, with three independent sampling seeds for all three backbones. We report $\mathrm{CHAIR}_{S}$ and $\mathrm{CHAIR}_{I}$ to measure hallucination at the caption and object-mention levels, respectively. Caption quality is assessed using BLEU-4~\cite{papineni2002bleu}, METEOR~\cite{banerjee2005meteor}, ROUGE-L~\cite{lin2004rouge}, and SPICE~\cite{anderson2016spice} to characterize retained content. For binary VQA, we evaluate the Random, Popular, and Adversarial settings of MS COCO POPE~\cite{li2023pope}, reporting accuracy and F1 for object-existence judgments. We additionally report perception, cognition, and total scores on the full MME benchmark~\cite{fu2025mme} to assess broader multimodal capabilities.

\paragraph{Baselines.}
We compare EGVCA with standard greedy decoding, beam search, and the native Yes--No decision, together with seven inference-time methods, VCD~\cite{leng2024vcd}, OPERA~\cite{huang2024opera}, DoLa~\cite{chuang2024dola}, DeCo~\cite{wang2025deco}, SID~\cite{huo2025sid}, MoLE~\cite{liang2025mole}, and HALC~\cite{chen2024halc}. These baselines allow us to compare claim admission with alternative inference-time interventions for improving factuality or mitigating hallucination without updating the evaluated LVLM. 

\paragraph{Implementation Details.}
For open-ended generation, EGVCA applies claim admission to greedy responses with $\tau=0.25$ for LLaVA-1.5, $\tau=0.30$ for MiniGPT-4, and $\tau=0.125$ for mPLUG-Owl2. For binary VQA, a single calibration parameter pair is selected for each backbone–benchmark setting using a predefined parameter sweep and is applied uniformly to all examples, using a held-out validation set. Appendix~\ref{sec:appendix_binary_sensitivity}
reports the local sensitivity around the selected operating points. CHAIR results are reported as the mean and standard deviation over three seeds, while caption-quality results are reported as three-seed means. All experiments were conducted on a single NVIDIA GeForce RTX 5090
GPU with 32 GB of memory.

\subsection{Experimental Results}
\label{sec:experimental_results}

\begin{table*}[t]
\centering
\setlength{\tabcolsep}{1.6pt}
\caption{POPE results under Random, Popular, and Adversarial sampling across three LVLM backbones.\tablefootnote{DoLa and HALC yielded
unstable binary outputs with MiniGPT-4 on both POPE and MME under
the matched protocol, indicating sensitivity to the
backbone--decoding interaction.}}
\label{tab:pope_results}
\resizebox{0.9\textwidth}{!}{%
\begin{tabular}{l*{18}{r}}
\toprule[1pt]
& \multicolumn{6}{c}{LLaVA-1.5}
& \multicolumn{6}{c}{MiniGPT-4}
& \multicolumn{6}{c}{mPLUG-Owl2} \\
\cmidrule(lr){2-7}
\cmidrule(lr){8-13}
\cmidrule(lr){14-19}
& \multicolumn{2}{c}{Random}
& \multicolumn{2}{c}{Popular}
& \multicolumn{2}{c}{Adversarial}
& \multicolumn{2}{c}{Random}
& \multicolumn{2}{c}{Popular}
& \multicolumn{2}{c}{Adversarial}
& \multicolumn{2}{c}{Random}
& \multicolumn{2}{c}{Popular}
& \multicolumn{2}{c}{Adversarial} \\
\cmidrule(lr){2-3}
\cmidrule(lr){4-5}
\cmidrule(lr){6-7}
\cmidrule(lr){8-9}
\cmidrule(lr){10-11}
\cmidrule(lr){12-13}
\cmidrule(lr){14-15}
\cmidrule(lr){16-17}
\cmidrule(lr){18-19}
Method
& Acc. & F1 & Acc. & F1 & Acc. & F1
& Acc. & F1 & Acc. & F1 & Acc. & F1
& Acc. & F1 & Acc. & F1 & Acc. & F1 \\
\midrule
Native
& 87.20 & 85.69 & 85.83 & 84.40 & \underline{83.47} & 82.25
& \underline{74.83} & \underline{72.56} & 71.40 & \underline{69.94}
& \underline{69.50} & \underline{68.70}
& \underline{88.67} & \underline{88.59}
& \underline{84.77} & \underline{85.24}
& 80.90 & \underline{82.12} \\

VCD
& 84.23 & 82.50 & 82.37 & 80.62 & 79.33 & 77.97
& 58.70 & 58.74 & 57.00 & 57.26 & 57.60 & 58.94
& 83.97 & 83.39 & 81.17 & 81.30 & 79.00 & 79.48 \\

OPERA
& 86.23 & 84.42 & 85.17 & 83.41 & 82.87 & 81.27
& 74.50 & 68.71 & \underline{72.03} & 66.69
& \textbf{69.67} & 65.00
& 86.83 & 85.80 & 84.40 & 83.60 & 80.80 & 80.51 \\

DoLa
& 88.43 & 87.44 & 86.37 & 85.52
& \textbf{83.57} & \underline{83.03}
& - & - & - & - & - & -
& 86.67 & 85.55 & 84.47 & 83.56
& \underline{81.07} & 80.61 \\

DeCo
& 88.53 & 87.60 & 86.33 & 85.56 & 83.37 & 82.95
& 52.57 & 67.78 & 51.77 & 67.42 & 51.83 & 67.45
& 86.47 & 86.68 & 82.30 & 83.27 & 78.03 & 80.04 \\

SID
& 89.40 & 89.05 & 86.10 & 86.11 & 80.47 & 81.49
& 59.07 & 57.74 & 58.13 & 57.07 & 56.57 & 55.39
& 84.03 & 83.60 & 80.70 & 80.87 & 77.10 & 77.99 \\

MoLE
& \underline{88.87} & \underline{88.11}
& \underline{86.87} & \underline{86.27}
& 83.20 & \textbf{83.05}
& 55.67 & 21.12 & 55.23 & 20.95 & 55.03 & 21.25
& 86.93 & 85.89 & 84.50 & 83.69 & 81.00 & 80.68 \\

HALC
& 88.50 & 87.52 & 86.43 & 85.60
& \textbf{83.57} & \textbf{83.05}
& - & - & - & - & - & -
& 86.53 & 85.85 & 83.87 & 83.51 & 80.20 & 80.46 \\

\midrule
EGVCA
& \textbf{90.43} & \textbf{90.05}
& \textbf{87.60} & \textbf{87.47}
& 80.97 & 81.94
& \textbf{76.80} & \textbf{78.45}
& \textbf{76.07} & \textbf{77.92}
& 69.40 & \textbf{73.41}
& \textbf{90.37} & \textbf{90.11}
& \textbf{87.37} & \textbf{87.42}
& \textbf{82.40} & \textbf{83.27} \\
\bottomrule[1pt]
\end{tabular}
}
\end{table*}

\begin{figure*}[t]
\centering
\includegraphics[width=0.8\textwidth]{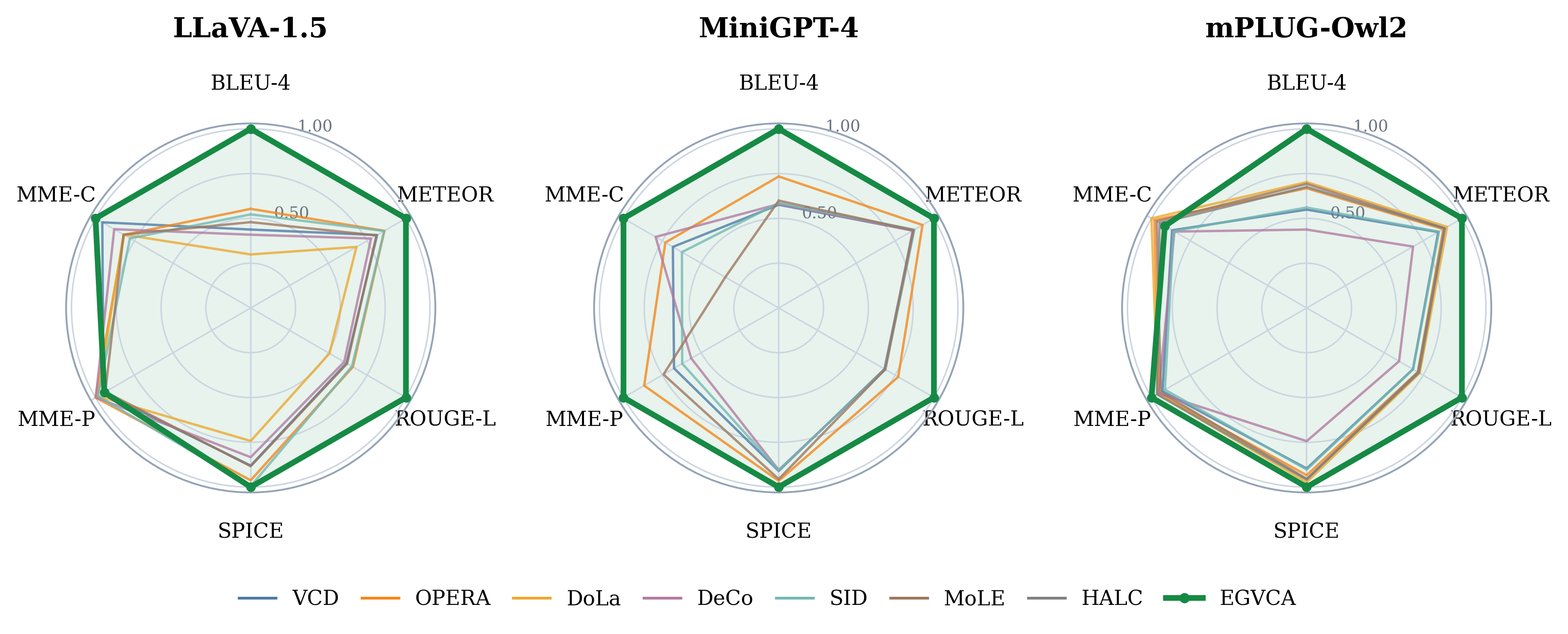}
\caption{General capability profiles across three LVLM backbones. BLEU-4, METEOR, ROUGE-L, and SPICE are evaluated on COCO captions,
whereas MME-P and MME-C are evaluated on the separate MME benchmark.
Each metric is normalized by the best matched result for the corresponding
backbone and metric. Greedy and Beam are omitted for clarity.}
\label{fig:general_capability_radar}
\end{figure*}

\paragraph{Open-ended Object Hallucination.}
Table~\ref{tab:chair_results}
 demonstrates the effectiveness of EGVCA in reducing object hallucination in open-ended generation. EGVCA achieves the lowest $\mathrm{CHAIR}_{S}$ in each backbone comparison. On LLaVA-1.5, it obtains a score of 10.87, outperforming DeCo, the strongest baseline on this metric, by 5.33 percentage points. On MiniGPT-4 and mPLUG-Owl2, EGVCA reduces $\mathrm{CHAIR}_{S}$ from 20.33 to 12.73 and from 21.47 to 12.60, respectively, while also improving $\mathrm{CHAIR}_{I}$. This result implies that a larger proportion of the resulting captions contain no hallucinated object mentions. This outcome aligns with the purpose of claim admission, which determines whether additional semantic content should enter the response. The improvements across all three backbones support the applicability of this principle to different LVLMs without modifying their underlying token-generation rules.

\paragraph{Binary Visual Question Answering.}
Table~\ref{tab:pope_results} shows that EGVCA improves object-presence prediction across different backbones and negative-sampling settings. EGVCA improves F1 over native predictions and achieves the highest
F1 among the evaluated methods in eight of the nine
backbone--setting combinations. Applied to LLaVA-1.5, EGVCA obtains the best accuracy and F1 in both the Random and Popular settings. On MiniGPT-4, it achieves the highest F1 across all three settings, with Popular F1 increasing from 69.94 to 77.92. On mPLUG-Owl2, EGVCA leads in both accuracy and F1 under every sampling setting. These results extend the evidence for claim admission beyond open-ended descriptions. Because POPE requires a binary response, its F1 improvements demonstrate that the benefit of EGVCA also applies to deciding whether an object-existence claim should be affirmed, beyond controlling how much descriptive content is generated.

\paragraph{General Capability Preservation.}
Figure~\ref{fig:general_capability_radar} jointly summarizes caption quality
and MME performance. For captioning, EGVCA achieves the best matched
BLEU-4, METEOR, ROUGE-L, and SPICE scores on all three backbones. On
LLaVA-1.5, SPICE increases from 21.44 with greedy decoding to 22.19, while
EGVCA obtains SPICE scores of 20.45 and 22.55 on MiniGPT-4 and
mPLUG-Owl2, respectively. The consistent improvements across all four
reference-based metrics indicate that hallucination reduction does not come
at the expense of semantic caption quality.

EGVCA also improves MME Total over the native model on all three
backbones, with the largest gain of 249.21 points on MiniGPT-4. It obtains
the highest cognition score on LLaVA-1.5, the best perception and cognition
scores on MiniGPT-4, and the highest perception and total scores on
mPLUG-Owl2. Together, these results show that EGVCA preserves broader
visual understanding while reducing hallucination.

\begin{table}[t]
\centering
\setlength{\tabcolsep}{3.5pt}
\caption{Sensitivity of caption admission to $\tau$ on LLaVA-1.5.
Results are the mean and standard deviation over three CHAIR-500 runs.}
\label{tab:caption_tau_ablation}
\resizebox{0.9\columnwidth}{!}{%
\begin{tabular}{lrrr}
\toprule[1pt]
Setting
& $\mathrm{CHAIR}_{S}\downarrow$
& $\mathrm{CHAIR}_{I}\downarrow$
& SPICE$\uparrow$ \\
\midrule
Greedy
& $21.93{\pm}1.79$
& $7.07{\pm}0.59$
& $21.44{\pm}0.30$ \\
$\tau=0$
& $20.47{\pm}1.48$
& $6.96{\pm}0.56$
& $21.67{\pm}0.25$ \\
$\tau=0.125$
& $14.73{\pm}1.47$
& $6.47{\pm}0.57$
& $\mathbf{22.23{\pm}0.14}$ \\
$\tau=0.25$ (Ours)
& $10.87{\pm}1.43$
& $5.77{\pm}0.61$
& \underline{$22.19{\pm}0.24$} \\
$\tau=0.5$
& \underline{$7.47{\pm}1.37$}
& \underline{$4.56{\pm}0.80$}
& $21.73{\pm}0.25$ \\
$\tau=1.0$
& $\mathbf{6.47{\pm}1.11}$
& $\mathbf{4.24{\pm}0.61}$
& $21.27{\pm}0.25$ \\
\bottomrule[1pt]
\end{tabular}}
\end{table}

\subsection{Ablation and Analysis.}
We analyze the key design choices of EGVCA on LLaVA-1.5. Specifically, we examine how the caption admission threshold affects the trade-off between hallucination reduction and semantic quality, assess the contributions of the visual-content and native-margin evidence components to binary decisions, and report the inference efficiency of the resulting framework.

\paragraph{Caption Admission Threshold.}
Table~\ref{tab:caption_tau_ablation} reports the sensitivity to
$\tau$ on CHAIR. Increasing $\tau$ consistently reduces both
$\mathrm{CHAIR}_{S}$ and $\mathrm{CHAIR}_{I}$, confirming that the
admission threshold directly controls the strictness of semantic
expansion. The best SPICE score is obtained at $\tau=0.125$, while
$\tau=0.25$ substantially reduces hallucination with only a negligible
SPICE difference (22.19 versus 22.23). We therefore select
$\tau=0.25$ as the final operating point. Relative to greedy
generation, it reduces $\mathrm{CHAIR}_{S}$ from 21.93 to 10.87 and
$\mathrm{CHAIR}_{I}$ from 7.07 to 5.77, while increasing SPICE from
21.44 to 22.19. More aggressive thresholds further reduce hallucination, but also decrease SPICE, demonstrating a trade-off between hallucination suppression and semantic quality.

\begin{table}[t]
\centering
\setlength{\tabcolsep}{3.5pt}
\caption{Binary component ablation on full POPE and MME using LLaVA-1.5.}
\label{tab:binary_component_ablation}
\resizebox{0.9\columnwidth}{!}{%
\begin{tabular}{lrrrr}
\toprule[1pt]
Variant
& POPE F1$\uparrow$
& MME-P$\uparrow$
& MME-C$\uparrow$
& MME Total$\uparrow$ \\
\midrule
Native LVLM
& 84.11
& 1360.14
& 321.79
& 1681.92 \\
w/o $M(q)$
& 81.93
& \underline{1371.71}
& \underline{361.79}
& \underline{1733.50} \\
w/o $V(q)$
& \underline{85.33}
& 1360.32
& 321.79
& 1682.11 \\
w/o $m_I(q)$
& 84.95
& 498.33
& 204.29
& 702.62 \\ \midrule
Full EGVCA
& \textbf{86.48}
& \textbf{1373.34}
& \textbf{366.79}
& \textbf{1740.13} \\
\bottomrule[1pt]
\end{tabular}}
\end{table}

\paragraph{Binary Evidence Components.}
Table~\ref{tab:binary_component_ablation} separates atomic visual
support $V(q)$ from native margin gain $M(q)$. Using $V(q)$ alone
substantially improves MME performance but degrades POPE, whereas $M(q)$ alone
improves POPE while leaving MME nearly unchanged. This indicates that
the two components capture complementary aspects of visual evidence.
Using the joint evidence as an independent classifier performs poorly,
particularly on MME, showing that counterfactual evidence should
calibrate rather than replace the LVLM's native decision. Combining
the native margin with joint evidence in the full admission head
achieves the strongest joint result, improving POPE F1 from 84.11 to
86.48 and MME Total from 1681.92 to 1740.13.

\begin{table}[t]
\centering
\small
\setlength{\tabcolsep}{5pt}
\caption{Inference efficiency on LLaVA-1.5 over CHAIR images.}
\label{tab:inference_efficiency}
\begin{tabular}{lrr}
\toprule[1pt]
Method
& Tokens/sec.$\uparrow$
& Peak VRAM (GB)$\downarrow$ \\
\midrule
Greedy
& \textbf{66.61}
& \textbf{15.27} \\
Beam
& 42.78
& 22.35 \\
VCD
& 33.56
& 16.37 \\
OPERA
& 5.43
& 25.45 \\
DoLa
& 19.69
& 26.31 \\
DeCo
& \underline{59.10}
& \underline{15.75} \\
SID
& 33.49
& 16.36 \\
MoLE
& 31.00
& 29.26 \\
HALC
& 1.45
& 19.54 \\
\midrule
EGVCA
& 21.90
& \textbf{15.27} \\
\bottomrule[1pt]
\end{tabular}
\end{table}

\paragraph{Inference Efficiency.}
Table~\ref{tab:inference_efficiency} reports end-to-end throughput and peak
GPU memory over 200 CHAIR images. EGVCA processes 21.90 final-output tokens
per second while using 15.27 GB of peak memory. Its throughput exceeds DoLa,
OPERA, and HALC, while its memory usage matches greedy decoding. Detailed
latency measurements and the EGVCA runtime breakdown are provided in the
appendix.

\section{Conclusion}

We presented EGVCA, a training-free framework that admits visual claims
according to their evidence under the observed image relative to a
visual-null input. Across three LVLM backbones, EGVCA
reduces CHAIR hallucination, improves POPE performance, and preserves or
improves caption quality and MME performance. These results demonstrate the
effectiveness of claim-level admission for controlling visual hallucination.

\section*{Limitations}
EGVCA requires additional original and visual-null scoring, resulting in
higher inference cost than standard greedy decoding. Our experiments focus
on three LVLM backbones and widely used English-language benchmarks.
Moreover, because EGVCA mitigates rather than verifies model outputs,
residual errors exist in the resulting responses. It should therefore not be treated as a factuality guarantee in high-stakes settings.




\bibliography{custom}

\clearpage

\appendix

\section{Additional Analyses}
\label{sec:appendix_additional_analyses}

\subsection{Hallucination across Claim Positions}
\label{sec:appendix_claim_position}

Previous studies report that LVLMs increasingly rely on linguistic context
as generation proceeds, while their dependence on visual information tends
to weaken in later decoding stages
\citep{min-etal-2025-mitigating,anand2026crops}. We examine whether the same
tendency appears across semantic claims in our setting. The analysis uses
the frozen LLaVA-1.5 Greedy captions from three CHAIR-500 runs, comprising
1,500 captions and 4,989 claims. We include all candidate claims, including
those occurring after the first operational rejection.

We report two complementary rates. The object-mention rate is the proportion
of recognized COCO object mentions labeled as hallucinated by the CHAIR
evaluator. The claim-level rate is the proportion of object-bearing claims
that contain at least one hallucinated object. Claims at position four and
later are pooled because their individual sample counts are smaller.

\begin{figure}[t]
\centering
\includegraphics[width=\columnwidth]
{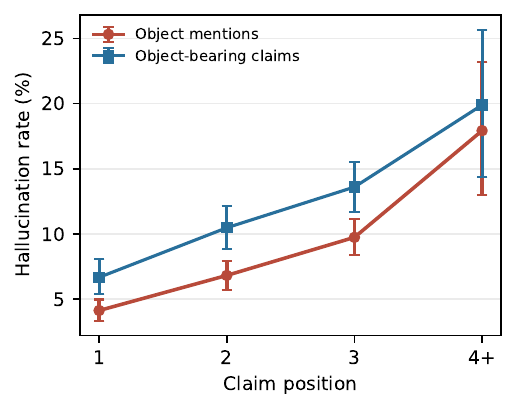}
\caption{Object hallucination across claim positions in LLaVA-1.5 Greedy
captions. Error bars indicate 95\% confidence intervals obtained using
image-grouped bootstrap resampling. Positions four and later are pooled.}
\label{fig:claim_position_hallucination}
\end{figure}

As shown in Figure~\ref{fig:claim_position_hallucination}, the
object-mention hallucination rate increases from 4.12\% in the first claim
to 6.81\% in the second, 9.74\% in the third, and 17.92\% at position four
or later. The proportion of object-bearing claims containing a hallucinated
object similarly increases from 6.67\% to 10.47\%, 13.60\%, and 19.90\%.
These results confirm that later claims are increasingly susceptible to
object hallucination in our generated captions, supporting the motivation
for controlling the admission of additional semantic content.

\subsection{Visual-Null Construction}
\label{sec:appendix_visual_null}

We compare EGVCA's zero tensor with VCD-style diffusion noise, a
CLIP-normalized black image, and a randomly shuffled COCO image. Captions
are always generated from the original image, and only the reference used
for evidence scoring is changed. Since diffusion noise can be sensitive to
its configuration \citep{Hyeonggeun2025AdjustingInitial}, we report it as a
separate reference condition.
As shown in Table~\ref{tab:appendix_visual_null}, 
the normalized-black
reference behaves similarly to the zero tensor, with slightly lower CHAIR
scores but slightly lower SPICE. In contrast, the diffusion-noise reference
is worse on both CHAIR metrics and SPICE and the shuffled-image reference produces
the highest hallucination rates despite maintaining a comparable SPICE score.
We use
the zero tensor because it retains the highest semantic quality and does not
require a stochastic corruption process.

\begin{table}[h]
\centering
\scriptsize
\setlength{\tabcolsep}{4pt}
\caption{Visual-null ablation on LLaVA-1.5 CHAIR-500.
Results are mean $\pm$ standard deviation over three seeds.}
\label{tab:appendix_visual_null}

\begin{tabularx}{\columnwidth}{
    @{}l
    *{3}{>{\centering\arraybackslash}X}
    @{}
}
\toprule[1pt]
Null reference
& $\mathrm{CHAIR}_{S}\downarrow$
& $\mathrm{CHAIR}_{I}\downarrow$
& SPICE$\uparrow$ \\
\midrule

Zero tensor (EGVCA)
& $\underline{10.87\pm1.43}$
& $\underline{5.77\pm0.61}$
& $\bm{22.19\pm0.24}$ \\

Normalized black
& $\bm{10.53\pm1.20}$
& $\bm{5.59\pm0.45}$
& $\underline{22.15\pm0.24}$ \\

VCD diffusion noise
& $11.13\pm2.07$
& $5.98\pm0.87$
& $22.04\pm0.24$ \\

Shuffled image
& $13.00\pm0.85$
& $6.63\pm0.38$
& $22.11\pm0.27$ \\

\bottomrule[1pt]
\end{tabularx}
\end{table}

\subsection{Binary Parameter Sensitivity}
\label{sec:appendix_binary_sensitivity}

Tables~\ref{tab:appendix_pope_sensitivity} and
\ref{tab:appendix_mme_sensitivity} present local $3\times3$ sensitivity
grids over $\tau$ and $\lambda$. The same decision equation,
$m'(q)=m_I(q)+\lambda[J(q)-\tau]$, is used throughout. On POPE, all
displayed settings outperform the native F1 score of 84.11, indicating
stable performance around the selected operating point. MME is more
sensitive to calibration, but eight of the nine settings remain above the
native Total score of 1681.92. In both benchmarks, the selected setting
achieves the best result within the reported local neighborhood.

\begin{table}[t]
\centering
\small
\setlength{\tabcolsep}{8pt}
\caption{Parameter sensitivity on POPE using LLaVA-1.5.
Rows vary $\tau$ and columns vary $\lambda$. Values are Avg.\ F1.}
\label{tab:appendix_pope_sensitivity}
\begin{tabular}{lrrr}
\toprule[1pt]
$\tau \backslash \lambda$
& 1.0 & 2.0 & 4.0 \\
\midrule
0.5
& 86.20 & \underline{86.42} & 85.22 \\
1.0
& 86.14 & \textbf{86.48} & 86.18 \\
1.5
& 85.89 & 85.65 & 86.07 \\
\bottomrule[1pt]
\end{tabular}
\end{table}

\begin{table}[t]
\centering
\small
\setlength{\tabcolsep}{8pt}
\caption{Parameter sensitivity on MME using LLaVA-1.5.
Rows vary $\tau$ and columns vary $\lambda$. Values are MME Total.}
\label{tab:appendix_mme_sensitivity}
\begin{tabular}{lrrr}
\toprule[1pt]
$\tau \backslash \lambda$
& 0.010 & 0.025 & 0.050 \\
\midrule
$-2.0$
& 1685.71 & \underline{1737.87} & 1706.80 \\
$-1.5$
& 1684.83 & \textbf{1740.13} & 1724.06 \\
$-1.0$
& 1678.08 & 1728.12 & 1737.34 \\
\bottomrule[1pt]
\end{tabular}
\end{table}

\subsection{Caption Admission Statistics}
\label{sec:appendix_admission_statistics}

We additionally inspect the decisions made by the caption admission gate on
the LLaVA-1.5 CHAIR-500 runs. Statistics are pooled over seeds 42, 777, and
2024, yielding 1,500 generated captions. The first claim is admitted
deterministically. Therefore, the admitted-evidence statistic in
Table~\ref{tab:appendix_admission_statistics} is computed only over later
claims. Admitted claims receive substantially stronger counterfactual evidence
than rejected claims. 


\begin{table}[!h]
\centering
\small
\setlength{\tabcolsep}{5pt}
\caption{Claim-admission statistics on LLaVA-1.5 CHAIR-500, pooled over three seeds.}
\label{tab:appendix_admission_statistics}
\begin{tabular}{lr}
\toprule[1pt]
Statistic & Value \\
\midrule
Number of captions & 1,500 \\
Candidate claims per caption & 3.326 \\
Admitted claims per caption & 1.462 \\
Captions stopped at a rejected claim & 92.27\% \\
Mean first-rejection position & 2.320 \\
Evidence of admitted post-first claims & $0.484\pm0.235$ \\
Evidence of rejected claims & $0.088\pm0.093$ \\
\bottomrule[1pt]
\end{tabular}
\end{table}


\subsection{Inference Efficiency}
\label{sec:appendix_efficiency}

We measure end-to-end inference efficiency on the same 200 CHAIR images
using LLaVA-1.5, batch size one, FP16 inference, and
\texttt{max\_new\_tokens}=64. We use three warm-up images followed by one
synchronized timing pass for each method. Model-loading time is excluded.
Because the methods generate outputs of different lengths, we report both
latency per image and final-output token throughput. For EGVCA, end-to-end
latency includes candidate generation, claim segmentation, original and
visual-null claim scoring, and claim admission.

\begin{table*}[t]
\centering
\setlength{\tabcolsep}{4pt}
\caption{Inference efficiency on LLaVA-1.5 over 200 CHAIR images after
three warm-up images. Each method is measured in one synchronized pass.}
\label{tab:appendix_efficiency}
\begin{tabular}{lrrrrr}
\toprule[1pt]
Method
& Sec./image$\downarrow$
& Tokens/sec.$\uparrow$
& Output tokens
& Peak VRAM (GB)$\downarrow$
& Relative overhead \\
\midrule
Greedy
& \textbf{0.953}
& \textbf{66.61}
& 63.48
& \textbf{15.27}
& \textbf{1.00$\times$} \\
Beam
& 1.493
& 42.78
& 63.88
& 22.35
& 1.57$\times$ \\
VCD
& 1.905
& 33.56
& 63.91
& 16.37
& 2.00$\times$ \\
OPERA
& 11.760
& 5.43
& 63.84
& 25.45
& 12.34$\times$ \\
DoLa
& 3.244
& 19.69
& 63.85
& 26.31
& 3.40$\times$ \\
DeCo
& \underline{1.080}
& \underline{59.10}
& 63.85
& \underline{15.75}
& \underline{1.13$\times$} \\
SID
& 1.905
& 33.49
& 63.80
& 16.36
& 2.00$\times$ \\
MoLE
& 2.058
& 31.00
& 63.81
& 29.26
& 2.16$\times$ \\
HALC
& 43.068
& 1.45
& 62.30
& 19.54
& 45.19$\times$ \\
\midrule
EGVCA
& 1.424
& 21.90
& 31.18
& \textbf{15.27}
& 1.49$\times$ \\
\bottomrule[1pt]
\end{tabular}
\end{table*}

EGVCA requires 1.49$\times$ the end-to-end latency of greedy decoding.
Its throughput exceeds DoLa, OPERA, and HALC, while its peak memory usage
matches that of greedy decoding. The lower final-output token count reflects
the shorter captions produced by claim admission. Table~\ref{tab:appendix_egvca_latency}
further separates the cost of candidate generation from the additional
claim-scoring stage.

\begin{table*}[t]
\centering
\setlength{\tabcolsep}{5pt}
\caption{End-to-end latency breakdown of EGVCA on LLaVA-1.5.}
\label{tab:appendix_egvca_latency}
\begin{tabular}{lr}
\toprule[1pt]
Component & Sec./image$\downarrow$ \\
\midrule
Candidate generation & 0.953060 \\
Claim segmentation & 0.000018 \\
Original/null claim scoring & 0.470771 \\
Admission decision & 0.000006 \\
\midrule
End-to-end & 1.423854 \\
\bottomrule[1pt]
\end{tabular}
\end{table*}

\subsection{Complete General-Capability Results}
Tables~\ref{tab:mme_results} and
\ref{tab:caption_quality_all} provide the complete numerical results
summarized by Figure~\ref{fig:general_capability_radar}. The dashes in Table~\ref{tab:mme_results} denote the omitted MiniGPT-4
results for DoLa and HALC, whose binary outputs were unstable under
the matched protocol.

\begin{table*}[t]
\centering

\setlength{\tabcolsep}{3pt}
\caption{General visual understanding on the full MME benchmark. Perception and cognition scores are reported alongside their total.}
\label{tab:mme_results}

\resizebox{0.9\textwidth}{!}{%
\begin{tabular}{lrrrrrrrrr}
\toprule[1pt]
& \multicolumn{3}{c}{LLaVA-1.5}
& \multicolumn{3}{c}{MiniGPT-4}
& \multicolumn{3}{c}{mPLUG-Owl2} \\
\cmidrule(lr){2-4}
\cmidrule(lr){5-7}
\cmidrule(lr){8-10}
Method
& Perception & Cognition & Total
& Perception & Cognition & Total
& Perception & Cognition & Total \\
\midrule
Native
& 1360.14 & 321.79 & 1681.92
& 693.94 & 196.79 & 890.73
& \underline{1398.51} & 329.64 & 1728.15 \\

VCD
& 1381.62 & \underline{350.71} & 1732.33
& 599.77 & 170.71 & 770.48
& 1310.22 & 331.43 & 1641.65 \\

OPERA
& \textbf{1456.40} & 299.29 & \underline{1755.68}
& \underline{771.00} & 182.86 & \underline{953.86}
& 1344.54 & \underline{375.00} & 1719.54 \\

DoLa
& 1434.32 & 300.71 & 1735.04
& - & - & -
& 1354.54 & \textbf{382.14} & \underline{1736.68} \\

DeCo
& \underline{1453.74} & 322.50 & \textbf{1776.24}
& 501.13 & \underline{198.57} & 699.70
& 1343.98 & 325.71 & 1669.69 \\

SID
& 1433.76 & 285.36 & 1719.11
& 552.97 & 156.07 & 709.04
& 1287.26 & 327.86 & 1615.11 \\

MoLE
& 1364.30 & 300.36 & 1664.65
& 660.29 & 86.07 & 746.36
& 1324.62 & 369.29 & 1693.91 \\

HALC
& 1440.07 & 295.36 & 1735.43
& - & - & -
& 1358.87 & 362.14 & 1721.01 \\

\midrule
EGVCA
& 1373.34 & \textbf{366.79} & 1740.13
& \textbf{889.58} & \textbf{250.36} & \textbf{1139.94}
& \textbf{1409.34} & 349.29 & \textbf{1758.62} \\
\bottomrule[1pt]
\end{tabular}}
\end{table*}

\begin{table*}[t]
\centering
\scriptsize
\setlength{\tabcolsep}{2.5pt}
\caption{Reference-based caption quality across three LVLM backbones.
Results are averaged over three seeds. }
\label{tab:caption_quality_all}
\resizebox{0.9\textwidth}{!}{%
\begin{tabular}{l*{12}{r}}
\toprule[1pt]
& \multicolumn{4}{c}{LLaVA-1.5}
& \multicolumn{4}{c}{MiniGPT-4}
& \multicolumn{4}{c}{mPLUG-Owl2} \\
\cmidrule(lr){2-5}
\cmidrule(lr){6-9}
\cmidrule(lr){10-13}
Method
& B-4$\uparrow$ & MET.$\uparrow$ & R-L$\uparrow$ & SPICE$\uparrow$
& B-4$\uparrow$ & MET.$\uparrow$ & R-L$\uparrow$ & SPICE$\uparrow$
& B-4$\uparrow$ & MET.$\uparrow$ & R-L$\uparrow$ & SPICE$\uparrow$ \\
\midrule
Greedy
& 7.89 & 23.01 & 27.01 & \underline{21.44}
& 7.94 & 22.52 & 26.37 & 19.07
& 8.05 & 23.26 & 27.25 & 21.72 \\

Beam
& \underline{8.04} & 23.11 & 27.55 & 21.26
& 8.30 & 23.09 & 26.89 & \underline{20.02}
& 8.48 & \underline{23.82} & 28.00 & 21.90 \\

VCD
& 6.36 & 21.92 & 25.88 & 19.53
& 7.17 & 22.02 & 26.03 & 18.61
& 6.65 & 22.31 & 26.37 & 20.21 \\

OPERA
& \underline{8.04} & \underline{23.19}
& \underline{27.63} & 21.35
& \underline{9.14} & \underline{23.54}
& \underline{29.40} & 19.74
& 8.09 & 23.17 & 27.59 & 21.06 \\

DoLa
& 4.34 & 18.35 & 21.29 & 16.48
& 5.08 & 17.93 & 21.62 & 14.99
& \underline{8.51} & \underline{23.82}
& \underline{28.03} & \underline{21.94} \\

DeCo
& 5.94 & 20.88 & 25.27 & 18.49
& 7.22 & 21.93 & 26.23 & 18.50
& 5.30 & 18.04 & 22.93 & 16.77 \\

SID
& 6.33 & 21.81 & 25.78 & 19.54
& 7.29 & 22.21 & 26.21 & 18.53
& 6.79 & 22.43 & 26.40 & 20.35 \\

MoLE
& 6.96 & 21.81 & 26.06 & 19.61
& 7.46 & 22.07 & 26.17 & 19.54
& 8.39 & 23.41 & 27.87 & 21.52 \\

HALC
& 4.47 & 20.06 & 23.64 & 17.92
& 6.49 & 21.05 & 25.41 & 17.86
& 8.18 & 23.39 & 27.61 & 21.60 \\

\midrule
EGVCA
& \textbf{14.53} & \textbf{26.93}
& \textbf{41.96} & \textbf{22.19}
& \textbf{12.45} & \textbf{25.41}
& \textbf{38.18} & \textbf{20.45}
& \textbf{12.11} & \textbf{26.33}
& \textbf{38.53} & \textbf{22.55} \\
\bottomrule[1pt]
\end{tabular}}
\end{table*}

\end{document}